\documentclass[a4paper,12pt]{article}

\usepackage[utf8]{inputenc}
\usepackage[T1]{fontenc}
\usepackage{amsmath}
\usepackage{amssymb}
\usepackage{amsthm}
\usepackage{amsfonts}
\usepackage{graphicx}
\usepackage[autolanguage,np]{numprint}
\usepackage{tikz}
\usepackage{xspace}
\usepackage{hyperref}
\usepackage{xcolor}
\usepackage{algorithm}
\usepackage{algorithmic}
\usepackage{mathtools}
\usepackage{booktabs}

\newcommand{\naturals}{\mathbb{N}}

\newcommand{\reals}{\mathbb{R}}

\newcommand{\binarySet}{\{0,1\}}
\newcommand{\hypercube}{\binarySet^n}

\newcommand{\norm}[1]{\left\|#1\right\|}
\newcommand{\range}[2]{[#1..#2]}

\DeclareMathOperator{\E}{\mathbb{E}}
\DeclareMathOperator{\proba}{\mathbb{P}}
\DeclareMathOperator*{\argmin}{arg\,min}
\DeclareMathOperator*{\argmax}{arg\,max}
\DeclareMathOperator*{\logit}{logit}

\begin{document}

\title{An information-theoretic evolutionary algorithm\footnote{A
    four-page version of this work will appear in \emph{Genetic and
      Evolutionary Computation Conference Companion (GECCO '23
      Companion)}. \url{https://doi.org/10.1145/3583133.3590548}}}

\author{Arnaud Berny}
\maketitle

\begin{abstract}
  We propose a novel evolutionary algorithm on bit vectors which
  derives from the principles of information theory. The
  information-theoretic evolutionary algorithm (it-EA) iteratively
  updates a search distribution with two parameters, the center, that
  is the bit vector at which standard bit mutation is applied, and the
  mutation rate. The mutation rate is updated by means of
  information-geometric optimization and the center is updated by
  means of a maximum likelihood principle. Standard elitist and non
  elitist updates of the center are also considered. Experiments
  illustrate the dynamics of the mutation rate and the influence of
  hyperparameters. In an empirical runtime analysis, on OneMax and
  LeadingOnes, the elitist and non elitist it-EAs obtain promising
  results.
\end{abstract}
 

\section{Introduction}

Information theory has revealed itself as a conceptual framework of
choice in the field of evolutionary computation, both in the analysis
of existing evolutionary algorithms and in the production of new ones
\footnote{Incidentally, the doctoral dissertation of Claude E.
  Shannon, the father of information theory, deals with theoretical
  genetics.}. This is especially true of estimation of distribution
algorithms \cite{larranaga2001estimation,hauschild2011introduction}
which sample an offspring population from a search distribution and
estimate its parameters using selected individuals. In particular,
algorithms such as MIMIC \cite{isbell} and UMDA \cite{muhlenbein-97}
completely rebuild a model at each iteration. They can be expressed in
terms of information geometry with a family of probability
distributions and a distance between distributions
\cite{toussaint2004notes,malago2008information}. Standard evolutionary
operators, such as mutation and cross-over, can also be interpreted in
this framework.

Information geometry is also a valuable resource for the analysis of
evolutionary algorithms such as PBIL \cite{baluja-caruana-95} and
CMA-ES \cite{hansen1996adapting,hansen2001completely} which estimate
an incremental change in the model parameters as opposed to a full
estimate. Starting from PBIL and CMA-ES, a strand of research has lead
to the formalization of this approach
\cite{berny2000selection,wierstra2008natural,malago2011towards,wierstra2014natural,ollivier2017information}.
In natural evolution strategies (NES)
\cite{wierstra2008natural,wierstra2014natural} and
information-geometric optimization (IGO)
\cite{ollivier2017information}, incremental changes are determined by
means of estimated natural gradients \cite{amari1998natural} with
respect to model parameters. The natural gradient is the gradient in
the Fisher metric of the search distribution which, in turn, is
expressed in terms of the Kullback–Leibler divergence
\cite{kullback1968information}. The key advantage of natural gradients
over Euclidean ones is that the evolution of the search distribution
does not depend on the choice of its parameters.

Throughout this paper, we will only be concerned with the search space
of fixed-length bit vectors. It is surprising that an evolutionary
algorithm on bit vectors similar to CMA-ES is missing. One may object
to this observation that PBIL is a natural candidate. However, its
parameters are all continuous and, in the form of a probability
vector, do not belong to the search space in the same way as, in
CMA-ES, the mean of the search distribution does. Furthermore, it is
expected that such a speculative evolutionary algorithm will be able
to control the mutation rate, just as CMA-ES provides an optimal
control of correlated mutations.

Adaptive parameter control in evolutionary algorithms has been the
subject of sustained efforts for decades
\cite{karafotias2015parameter,aleti2016systematic}. It has been shown,
both in practice and in theory \cite{doerr2020theory}, to provide an
advantage over fixed parameter strategies. However, update rules are
often introduced as heuristics and do not derive from first
principles.

A famous update rule is the so called one-fifth success rule
\cite{rechenberg1973evolutionsstrategie} which adjusts the mutation
rate so as to keep the success rate of mutations equal or close to one
fifth (a mutation is successful if it produces an offspring of
increased fitness). If the success rate is larger (smaller) than one
fifth then the mutation rate should be increased (decreased). The
one-fifth constant can be derived from theoretical considerations
about the $(1+1)$ evolution strategy applied to the sphere problem in
Euclidean spaces. It has been proved that the $(1+1)$ evolutionary
algorithm equipped with a similar success-based update rule achieves
the same performance on LeadingOnes as the $(1+1)$ EA with an optimal
fitness-dependent mutation rate \cite{doerr2021self}. The 2-rate
$(1+\lambda)$ EA with success-based mutation rates achieves the best
performance on OneMax among all $\lambda$-parallel black-box
algorithms \cite{doerr2017theone}.

In another line of research, the problems of operator selection and
parameter control have been addressed by means of machine learning,
including adaptive pursuit \cite{thierens2005adaptive} and dynamic
multi-armed bandits \cite{dacosta2008adaptive}, or reinforcement
learning \cite{karafotias2014generic}. These methods often assume a
finite set of operators or values, hence require the discretization of
continuous parameters such as the mutation rate. It is also possible
to reason about the number of mutated bits \cite{doerr2016kbit}
instead of the mutation rate itself. Finally, self-adaptation has been
applied to control the mutation rate in genetic algorithms as in
evolution strategies \cite{back1992interaction}.

In this work, we propose a novel evolutionary algorithm on bit vectors
which derives from the principles of information theory. The
information-theoretic evolutionary algorithm (it-EA) iteratively
updates a search distribution with two parameters, the center, that is
the bit vector at which standard bit mutation is applied, and the
mutation rate. The mutation rate is updated by means of
information-geometric optimization \cite{ollivier2017information} and
the center is updated by means of a maximum likelihood (ML) principle,
similar to the cross-entropy method (CEM)
\cite{rubinstein1999cross,deboer2005tutorial}. Standard elitist and
non elitist updates of the center are also considered.

The paper is organized as follows. The mutation model is presented in
Sect.~\ref{sec:mutation-model}. The selection model is presented in
Sect.~\ref{sec:selection-model}. In Sect.~\ref{sec:igo-flow} and
\ref{sec:igo-update}, building upon the IGO framework, we derive a
precise update rule for the mutation rate. In
Sect.~\ref{sec:replacement}, we apply a maximum likelihood principle
to derive an update rule for the center of the search distribution.
The information-theoretic evolutionary algorithm which embeds all the
previous principles is presented in Sect.~\ref{sec:evol-algor}.
Sect.~\ref{sec:experiments} provides experiments with the it-EA.
Sect.~\ref{sec:conclusion} concludes the paper.

\section{Mutation model}
\label{sec:mutation-model}

We recall the model of standard mutation of bit vectors and introduce
notations as needed. We consider the set $\hypercube$ of
$n$-dimensional bit vectors, where $n\in\naturals^*$. The model of
standard bit mutation is given by a random bit vector $U$ with
independent and identically distributed components. More precisely,
for all $u\in\hypercube$,
\begin{equation*}
  \proba(U = u) = \prod_i \proba(U_i = u_i) \,,
\end{equation*}
where $U=(U_i)$, $u=(u_i)$, and $i\in\range{1}{n}$. Let $p\in(0,1)$ be
the mutation rate. Then,
\begin{equation*}
  \proba(U_i = u_i) =
  \begin{cases}
    p & \text{if $u_i = 1$} \\
    1 - p & \text{if $u_i = 0$}
  \end{cases}
\end{equation*}
and
\begin{equation*}
  \proba(U = u) = p^{\norm{u}} (1 - p)^{n - \norm{u}} \,,
\end{equation*}
where $\norm{u} = \sum_i u_i$ is the Hamming weight (bit count) of
$u$. We will denote by $Q_p$ the probability distribution of $U$, that
is $Q_p(u) = \proba(U = u)$.

Let $x$ be some fixed bit vector and suppose that we want to explore
the neighborhood of $x$. We define the random vector $V = x \oplus U$,
where $U$ has the probability distribution $Q_p$ ($U \sim Q_p$ for
short) and $\oplus$ denotes addition modulo 2. This is similar to the
Gaussian case, where $U$ would be a centered Gaussian random vector
and $x$ the mean of $V$. However, in the case of bit vectors and
addition modulo 2, we cannot speak of mean. Instead, we will say that
$x$ is the center of the search distribution.

It turns out that the set $\{Q_p : p\in(0, 1)\}$ defines an
exponential family of probability distributions
\cite{malago2011towards,ollivier2017information}. Indeed, for all
$u\in\hypercube$, $Q_p(u) = P_\theta(u)$ with
\begin{equation}
  \label{eq:1}
  P_\theta(u) = \frac1{Z(\theta)} e^{\theta T(u)} \,,
\end{equation}
where $\theta = \logit(p)$,
\begin{equation*}
  \logit(p) = \log\left(\frac{p}{1-p}\right) \,,
\end{equation*}
$T(u) = \norm{u}$, and $Z(\theta) = (1+e^\theta)^n$ is a normalizing
constant. The reasons to choose $P_\theta$ over $Q_p$ are that the IGO
flow has a simple expression with exponential families and the
evolution of the search distribution is invariant under
reparametrization.

\section{Selection model}
\label{sec:selection-model}

In a black-box context, we want to maximize an arbitrary function
$f : \hypercube \rightarrow \reals$ defined on bit vectors of
dimension $n \in \naturals^*$. Let $x\in\hypercube$ be some fixed bit
vector. Let us define $g$, the translated version of $f$, by
$g(u) = f(x \oplus u)$, for all $u\in\hypercube$. As in the previous
section, $u$ models mutated bits. Following the IGO framework
\cite{ollivier2017information}, we replace $g$ with a selection weight
$W_\theta^g$. Roughly speaking, for a given $u\in\hypercube$, the
smaller the probability of improvement over $u$, the larger its
selection weight $W_\theta^g(u)$.

Assuming maximization of $f$ hence $g$, we define two probabilities of
improvement under the sampling distribution $P_\theta$. More
precisely, for all $u\in\hypercube$,
\begin{align}
  q^{>}(u) &= \proba(g(V) > g(u)) \label{eq:6} \\
  q^{\ge}(u) &= \proba(g(V) \ge g(u)) \notag \,,
\end{align}
where $V \sim P_\theta$. The distinction between comparison operators
is necessary since the support of $P_\theta$ is $\hypercube$, which
implies that
\begin{equation*}
  \proba(g(V) = g(u)) = q^{\ge}(u) - q^{>}(u)
\end{equation*}
is positive for all $u\in\hypercube$. Without this distinction, we
could define the selection weight, for all $u\in\hypercube$, by
$W_\theta^g(u) = w(q(u))$, where $w : [0, 1] \rightarrow \reals$ is a
non-increasing function. Instead, it is defined as the mean value of
$w$ on the interval $[q^{>}(u), q^{\ge}(u)]$ or
\begin{equation}
  W_\theta^g(u) = \frac{1}{q^{\ge}(u) - q^{>}(u)}
  \int_{q^{>}(u)}^{q^{\ge}(u)} w(q) \, dq \,.
  \label{eq:2}
\end{equation}
An important property of the selection weight is that, for all
$\theta\in\reals$,
\begin{equation*}
  \E_{u \sim P_\theta} (W_{\theta}^g(u)) = \int_0^1 w(q) \, dq \,,
\end{equation*}
which only depends on $w$. We will choose $w$ so as to make this
expectation equal to 1.

In CMA-ES, at each iteration, $\lambda$ individuals are sampled from
$P_\theta$ and the $\mu < \lambda$ best ones, according to the fitness
function, are used to update the parameter $\theta$. This behavior can
be modelled in IGO with the step function $w$ defined, for all
$q\in[0, 1]$, by
\begin{equation}
  \label{eq:7}
  w(q) = \frac1{q_0} 1_{q\le q_0} \,,
\end{equation}
where $q_0 = \mu / \lambda$.

\section{IGO flow}
\label{sec:igo-flow}

In the IGO framework, we replace the function to maximize by the
expectation of the selection weight with respect to the search
distribution. More precisely, let us define the 2-variable function
\begin{equation*}
  \varphi(\theta, \theta') = \E_{u \sim P_\theta} (W_{\theta'}^g(u)) \,.
\end{equation*}
Then, the IGO flow acting on $\theta$ is defined by
\begin{equation*}
  \frac{d\theta_t}{dt} = \tilde{\nabla}_\theta \varphi(\theta,
  \theta_t) \,,
\end{equation*}
where the natural gradient, denoted by $\tilde{\nabla}$, is taken at
$\theta = \theta_t$. The natural gradient, as opposed to the Euclidean
gradient, is necessary to ensure that the resulting flow does not
depend on the particular parametrization that we have chosen for the
probability distribution $P_\theta$. Although $\theta$ is a scalar, we
will stick with the gradient nomenclature. We can express the gradient
of the expectation as the expectation of another gradient
(log-derivative trick) and write
\begin{equation*}
  \frac{d\theta_t}{dt} = \E_{u \sim P_{\theta_t}} \left( W_{\theta_t}^g(u) \cdot \tilde{\nabla}_\theta \log P_\theta(u) \right) \,,
\end{equation*}
where, again, the natural gradient is taken at $\theta = \theta_t$.
Instead of explicitly computing the natural gradient, we take
advantage of the fact that the mutation model defines an exponential
family of probability distributions, as we have seen in
Sect.~\ref{sec:mutation-model}. The expression of the log-derivative
can be found in \cite{malago2011towards,ollivier2017information}. The
IGO flow can then be expressed as
\begin{equation*}
  \frac{d\overline{T}}{dt} = \E_{u \sim P_{\theta_t}}
  \left(W_{\theta_t}^g(u) \cdot (T(u) - \overline{T})\right) \,,
\end{equation*}
where $\overline{T} = \E_{u \sim P_{\theta_t}}(T(u))$. Since
$T(u) = \norm{u}$ and $u \sim P_{\theta_t}$, we have
$\norm{u} \sim \mathcal{B}(n, p_t)$ (binomial distribution) and
$\overline{T} = np_t$. Dividing by $n$ and recalling that
$\E_{u \sim P_{\theta_t}} (W_{\theta_t}^g(u)) = 1$, we obtain
\begin{align}
  \frac{dp_t}{dt} &= \E_{u \sim P_{\theta_t}}
                    \left(W_{\theta_t}^g(u) \cdot \left(\frac{\norm{u}}{n} - p_t\right)\right) \notag \\
                  &= \E_{u \sim P_{\theta_t}}
                    \left(W_{\theta_t}^g(u) \cdot \frac{\norm{u}}{n}\right) - p_t \,.
                    \label{eq:3}
\end{align}
The flow can now be interpreted as follows. If the frequency of
mutated bits in selected individuals is greater (resp.\ lower) than
the current mutation rate then increase (resp.\ decrease) it. This is
similar to the behavior of the IGO flow acting on the parameters of a
Gaussian distribution, in particular its standard deviation. An
asymptotic value $p$ for the dynamical system~(\ref{eq:3}) must verify
\begin{equation}
  \label{eq:10}
  p = \E_{u \sim P_{\theta}} \left(W_{\theta}^g(u) \cdot
    \frac{\norm{u}}{n}\right) \,,
\end{equation}
which means that $p$ must be equal to the frequency of mutated bits in
selected individuals. There can be several such fixed-points even if
$p_t$ can only converge to a unique asymptotic value given an initial
condition $p_0$.

\section{IGO update}
\label{sec:igo-update}

Our aim is to provide an evolutionary algorithm with adaptive mutation
rate by taking advantage of the theory exposed in the previous
section. Ideally, we would like to compute the asymptotic value
defined by Eq.~(\ref{eq:10}) given an initial condition $p_0$. Due to
computational limits, we will use an approximation of
Eq.~(\ref{eq:3}), called an IGO update \cite{ollivier2017information},
where time is discretized and expectations over $\hypercube$ are
estimated by sums of random variables. Even then, we will only afford
a single update of the mutation rate per iteration of the evolutionary
algorithm.

Let us describe the IGO update starting from the expectation in
Eq.~(\ref{eq:3}) at discrete time $t$. As in previous sections, let
$x\in\hypercube$ be some fixed bit vector and define $g$ by
$g(u) = f(x \oplus u)$, for all $u\in\hypercube$. First, we sample
$\lambda$ mutation bit vectors $u^k \sim Q_{p_t}$,
$k\in\range{1}{\lambda}$, and compute their respective images under
$g$. Let $\widehat{W}_k$ be the approximation of the selection weight
$W_{\theta_t}^g(u^k)$ of $u^k$. We will precisely define it later.
Then, the expectation is estimated by
\begin{equation}
  \label{eq:8}
  \widehat{p} = \frac1{\lambda} \sum_{k=1}^\lambda \widehat{W}_k \cdot \frac{\norm{u^k}}{n}
\end{equation}
and the mutation rate is updated with
\begin{equation}
  \label{eq:9}
  p_{t+1} = p_t + \alpha \cdot (\widehat{p} - p_t) \,,
\end{equation}
where $\alpha$ is the learning rate or time increment. By Theorem~6 in
\cite{ollivier2017information}, when $\lambda \rightarrow \infty$, the
estimator $\widehat{p}$ converges with probability 1 to the
expectation in the RHS of Eq.~(\ref{eq:3}) (consistency).

We now turn to the selection weights. We need to define ranks in the
same way as improvement probabilities in Eq.~(\ref{eq:6}). For all
$k\in\range{1}{\lambda}$, let
\begin{align*}
  r_k^> &= \# \{ \ell \in\range{1}{\lambda} : g(u^\ell) > g(u^k) \} \text{ and } \\
  r_k^\ge &= \# \{ \ell \in\range{1}{\lambda} : g(u^\ell) \ge g(u^k) \} \,.
\end{align*}
Then $r_k^> / \lambda$ is an approximation of $q^{>}(u^k)$ defined in
Eq.~(\ref{eq:6}). We replace the integral in Eq.~(\ref{eq:2}) with a
finite sum and define the empirical weights
\begin{equation}
  \widehat{W}_k = \frac1{r_k^\ge - r_k^>} \sum_{\ell = r_k^>}^{r_k^\ge-1}
  w\left(\frac{\ell + 1/2}{\lambda}\right) \,,
  \label{eq:4}
\end{equation}
where $w$ is the step function defined in Eq.~(\ref{eq:7}). We have to
take care of equivalent individuals in the computation of the weights.
This is particularly important in the case of functions defined on
$\hypercube$. Very often such functions take a small number of values
compared to the size of the search space. Therefore, we divide the set
$\range{1}{\lambda}$ into three pairwise disjoint and possibly empty
subsets depending on the position of $\mu$ relatively to the real
interval $[r_k^> + 1/2, r_k^\ge - 1/2]$.
\begin{itemize}
\item Let $K_1 = \{ k\in\range{1}{\lambda} : \mu \ge r_k^\ge \}$.
  Then, for all $k\in K_1$, $\widehat{W}_k = \lambda / \mu$. The
  computation of the weights is a direct consequence of the definition
  of the step function $w$ and will be omitted.
\item Let
  $K_2 = \{ k\in\range{1}{\lambda} : r_k^> + 1 \le \mu \le r_k^\ge - 1
  \}$. Then, for all $k\in K_2$, the sum in the right-hand side of
  Eq.~(\ref{eq:4}) is truncated and
  \begin{equation*}
    \widehat{W}_k
    = \frac1{r_k^\ge - r_k^>} \sum_{\ell = r_k^>}^{\mu-1} w\left(\frac{\ell + 1/2}{\lambda}\right)
    = c \cdot \frac{\lambda}{\mu} \,,
  \end{equation*}
  where
  \begin{equation*}
    c = \frac{\mu - r_k^>}{r_k^\ge - r_k^>} \,.
  \end{equation*}
  An interpretation of the coefficient $c$ is as follows. The
  $\mu - r_k^>$ weights $\lambda / \mu$ are evenly distributed across
  all $r_k^\ge - r_k^>$ equivalent individuals. For example, if there
  are two equivalent individuals, each one of them will be assigned a
  weight equal to $\lambda / (2\mu)$. This is in contrast to the
  simpler rule according to which one individual would be assigned a
  weight equal to $\lambda / \mu$ whereas the other one would be
  assigned a weight equal to 0.
\item Let $K_3 = \{ k\in\range{1}{\lambda} : \mu \le r_k^> \}$. Then,
  for all $k\in K_3$, $\widehat{W}_k = 0$.
\end{itemize}

From the definitions of the subsets, assuming none is empty, it can be
verified that
\begin{align*}
  \sum_{k=1}^\lambda \widehat{W}_k
  &= \sum_{k\in K_1} \widehat{W}_k + \sum_{k\in K_2} \widehat{W}_k +
    \sum_{k\in K_3} \widehat{W}_k \\
  &= r_\ell^> \cdot \frac{\lambda}{\mu}
    + (r_\ell^\ge - r_\ell^>) \cdot \frac{\mu - r_\ell^>}{r_\ell^\ge - r_\ell^>} \cdot \frac{\lambda}{\mu}
    + 0 = \lambda \,,
\end{align*}
where $\ell$ has been arbitrarily chosen in $K_2$. This is consistent
with the fact that $\E_{u \sim P_\theta} (W_{\theta}^g(u)) = 1$, as
emphasized in Sect.~\ref{sec:selection-model}.

Finally, it should be noted that those selection weights are also
relevant to other algorithms, such as PBIL, UMDA, or even CMA-ES when
applied to functions taking discrete values.

\section{ML update}
\label{sec:replacement}

In the previous sections, we have been concerned with the control of
the mutation rate. We now consider ways of updating the center of the
search distribution in the it-EA. Beside elitist and non elitist
selections found in the $(1+\lambda)$ and $(1, \lambda)$ evolutionary
algorithms, we propose to apply information theory to this problem as
well. More precisely, we apply the cross-entropy method (CEM)
\cite{rubinstein1999cross,deboer2005tutorial}. For the relation
between IGO and CEM see \cite{ollivier2017information}.

The main idea is to update the center $x_t$, a discrete parameter of
the search distribution, so as to minimize the Kullback–Leibler
divergence \cite{kullback1968information} between selected individuals
and the model. Put another way,
\begin{equation*}
  x_{t+1} = \argmin_y D_\text{KL}
  (P'_{x_t, \theta_t} \parallel P_{y, \theta_t}) \,,
\end{equation*}
where $P_{x_t, \theta_t}(x) = P_{\theta_t}(x_t \oplus x)$ is the
probability distribution $P_{\theta_t}$ translated by $x_t$ and
$P'_{x_t, \theta_t} = W_{\theta_t}^f \cdot P_{x_t, \theta_t}$ is its
image under selection. In what follows, we directly use $f$ instead of
$g$ as in Sect.~\ref{sec:selection-model} to \ref{sec:igo-update}.
Since $\E_{x \sim P_{x_t, \theta_t}} (W_{\theta_t}^f(x)) = 1$,
$P'_{x_t, \theta_t}$ is indeed a probability distribution.

Up to a constant with respect to $y$, the KL divergence is equal to
the cross-entropy between $P'_{x_t, \theta_t}$ and $P_{y, \theta_t}$
defined by
\begin{equation*}
  H(P'_{x_t, \theta_t}, P_{y, \theta_t})
  = -\sum_x P'_{x_t, \theta_t}(x) \log P_{y, \theta_t}(x) \,.
\end{equation*}
The cross-entropy is approximated in the same way as the expectation
in Eq.~(\ref{eq:3}) by
\begin{equation*}
  \widehat{H}(y) = -\frac{1}{\lambda}
  \sum_{k=1}^\lambda \widehat{W}_k \log P_{y, \theta_t}(x^k) \,.
\end{equation*}
By Theorem~6 in \cite{ollivier2017information}, when
$\lambda \rightarrow \infty$, the estimator $\widehat{H}(y)$ converges
with probability 1 to the cross-entropy (consistency). The function
$\widehat{H}$ is the opposite of a log-likelihood. We will see that it
can be exactly minimized. By the definitions of $P_{y, \theta_t}$ and
$P_{\theta_t}$,
\begin{align*}
  \log P_{y, \theta_t}(x^k)
  &= \log P_{\theta_t}(y \oplus x^k) \\
  &= \norm{y \oplus x^k} \cdot \log \frac{p_t}{1 - p_t} + n \log(1 -
    p_t) \,.
\end{align*}
Up to a constant with respect to $y$,
\begin{align*}
  \widehat{H}(y)
  &= \frac{1}{\lambda} \sum_{k=1}^\lambda \widehat{W}_k \cdot \norm{y
    \oplus x^k} \cdot \log \frac{1 - p_t}{p_t} \\
  &= \sum_{i = 1}^n \mathcal{L}_i(y_i) \cdot \log \frac{1 - p_t}{p_t} \,,
\end{align*}
which is a completely separable function, where
\begin{equation*}
  \mathcal{L}_i(y_i) = \frac{1}{\lambda} \sum_{k=1}^\lambda
  \widehat{W}_k \cdot (y_i \oplus x_i^k) \,.
\end{equation*}
The logarithmic factor does not depend on $y$ and only influences the
direction of the optimization. For a given $i\in\range{1}{n}$, up to a
constant with respect to $y_i$,
\begin{align*}
  \mathcal{L}_i(y_i)
  &= \frac{1}{\lambda} \sum_{k : x_i^k = 1} \widehat{W}_k \cdot (1 - y_i) +
    \frac{1}{\lambda} \sum_{k : x_i^k = 0} \widehat{W}_k \cdot y_i \\
  &= \left(1 - \frac{2}{\lambda} \sum_{k : x_i^k = 1} \widehat{W}_k \right) y_i \,.
\end{align*}
We can now state the update rule. Assume $p_t\in(0, 1/2)$ or,
equivalently, $\log \frac{1 - p_t}{p_t} > 0$. Then
\begin{equation}
  \label{eq:5}
  x_{i, t+1} =
  \begin{cases}
    0 & \text{if $\frac{1}{\lambda} \sum_{k : x_i^k = 1} \widehat{W}_k < \frac{1}{2}$,} \\
    1 & \text{if $\frac{1}{\lambda} \sum_{k : x_i^k = 1} \widehat{W}_k > \frac{1}{2}$.}
  \end{cases}
\end{equation}
In case of equality, $x_{i, t+1}$ is uniformly sampled in
$\binarySet$. Therefore, $x_{i, t+1}$ is set by means of a majority
rule. In the case $p\in(1/2, 1)$, we have to reverse the inequality
signs.

We will refer to this update rule as the ML update. The ML update can
be seen as a multi-parent cross-over operator as it combines all
selected individuals to form the next center. On one hand it favors
intensification since its purpose is to increase the probability of
selected individuals. On the other hand it also favors exploration to
some degree since the update rule can deteriorate the fitness of the
center, in contrast to local search. Moreover, the Hamming distance
between the old and the new center is not upper bounded by 1. Finally,
if $\mu = 1$, ignoring equivalent individuals, the ML update is
similar to the non elitist $(1, \lambda)$ selection for replacement.

We will also consider a local version of the ML update defined by
\begin{equation*}
  x_{t+1} = x_t \oplus \argmin_{\norm{u}\le 1}
  D_\text{KL}
  (P'_{x_t, \theta_t} \parallel P_{x_t\oplus u, \theta_t}) \,.
\end{equation*}
It leads to an update rule similar to Eq.~(\ref{eq:5}) where
individuals $x^k$ are replaced by mutations $u^k$. The resulting
incremental ML update can be interpreted as an incremental
multi-parent cross-over.

\section{Evolutionary algorithm}
\label{sec:evol-algor}

The proposed information-theoretic evolutionary algorithm (it-EA, see
Algorithm~\ref{algo:it_ea}) combines the IGO update (see
Sect.~\ref{sec:igo-update}) and the ML update (see
Sect.~\ref{sec:replacement}). It takes five parameters: the fitness
function $f$, the number $\lambda$ of sampled individuals, the number
$\mu$ of selected individuals, the initial mutation rate
$p_0\in(0,1)$, and the learning rate $\alpha\in[0,1]$.

The it-EA is an iterative algorithm which updates its state made of
the center $x$ of the search distribution and the mutation rate $p$.
At the beginning of each iteration, the algorithm samples a population
of bit vectors, much like a standard $(1+\lambda)$ or $(1,\lambda)$
EA. The population is evaluated and sorted. The algorithm determines
the set $K_2$ and the coefficient $c$ to take into account equivalent
individuals (see Sect.~\ref{sec:igo-update}). Finally, the center of
the search distribution and the mutation rate are updated. The scaled
mutation rate $np$ (the expected number of mutated bits) can be seen
as the radius of a random local search which takes place at the
center. From this perspective, the it-EA implements a form of variable
neighborhood local search.

We introduce two binary relations on $\range{1}{\lambda}$ to
facilitate both the reading of the algorithm and the reasoning about
it. They depend on a population $(x^k)$, $k\in\range{1}{\lambda}$, of
bit vectors. The first relation is the preorder defined by
$k\preceq \ell$ if $f(x^k) \ge f(x^\ell)$. It is used to sort the
range $\range{1}{\lambda}$. The sort function returns a permutation
$\sigma$ of $\range{1}{\lambda}$. For example, $\sigma(1)$ is the
index of the fittest individual. The second relation is the
equivalence relation defined by $k\equiv \ell$ if
$f(x^k) = f(x^\ell)$. The equivalence class of $k$ will be denoted by
$[k]$. It can be expressed as the image of the integer interval
$\range{r_k^> + 1}{r_k^\ge}$ under $\sigma$, where we are using the
ranks defined in Sect.~\ref{sec:igo-update}. That is
$[k] = \sigma(\range{r_k^> + 1}{r_k^\ge})$.

The set $K_2$ is not empty if the equivalence class $[\sigma(\mu)]$ of
the $\mu$-th individual extends beyond the first $\mu$ individuals
$\sigma(\range{1}{\mu})$ or, equivalently, if
$\sigma(\mu) \equiv \sigma(\mu+1)$. In this case,
$K_2 = [\sigma(\mu)]$,
$K_1 = \sigma(\range{1}{\mu}) \smallsetminus K_2$, and
\begin{equation*}
  c
  = \frac{\mu - r_\ell^>}{r_\ell^\ge - r_\ell^>}
  = \frac{\# \sigma(\range{1}{\mu}) \cap K_2}{\# K_2} \,,
\end{equation*}
where $\ell$ has been arbitrarily chosen in $K_2$.

The it-EA has four hyperparameters, $\lambda$, $\mu$, $p_0$, and
$\alpha$, that is two more than the standard $(1+\lambda)$ EA. The
update rule for the mutation rate ensures $0\le p\le 1$ but, if the
value of the mutation rate becomes too small, the search nearly comes
to a halt. To avoid this situation, we can set bounds on the values of
the mutation rate with the line
$p\leftarrow \min(p_{\max}, \max(p_{\min}, p))$ at the end of the main
loop so that $p\in[p_{\min}, p_{\max}]$.

We can easily turn Algorithm~\ref{algo:it_ea} into an elitist
$(1 + \lambda)$ EA with the line
$x \leftarrow \argmax_{\{x, x^1, \ldots, x^\lambda\}} f$ or into a
non-elitist $(1,\lambda)$ EA with the line
$x \leftarrow \argmax_{\{x^1, \ldots, x^\lambda\}} f$ in place of the
ML update. The elitist and non elitist it-EAs will be denoted by
eit-EA and neit-EA respectively. The it-EA with the incremental ML
update which flips at most one bit (see Sect.~\ref{sec:replacement})
will be denoted by it1-EA.

\begin{algorithm}
  \caption{Information-theoretic evolutionary algorithm.}
  \label{algo:it_ea}
  \begin{algorithmic}[1]
    \STATE Input parameters: $(f, \lambda, \mu, p_0, \alpha)$
    \STATE Sample $x$ uniformly on $\hypercube$
    \STATE Evaluate $f(x)$
    \STATE $p \leftarrow p_0$
    \LOOP
    \FORALL{$k\in\range{1}{\lambda}$}
    \STATE Sample $u^k \sim Q_p$
    \STATE $x^k \leftarrow x \oplus u^k$
    \STATE Evaluate $f(x^k)$
    \ENDFOR
    \STATE $\sigma\leftarrow \text{sort}(\range{1}{\lambda}, \preceq)$
    \IF{$\sigma(\mu) \equiv \sigma(\mu+1)$}
    \STATE $K_2 \leftarrow [\sigma(\mu)]$
    \STATE $c \leftarrow \frac{\# \sigma(\range{1}{\mu}) \cap K_2}{\# K_2}$
    \ELSE
    \STATE $K_2 \leftarrow \emptyset$
    \STATE $c\leftarrow 0$
    \ENDIF
    \STATE $K_1 \leftarrow \sigma(\range{1}{\mu}) \smallsetminus K_2$
    \FORALL{$i\in\range{1}{n}$}
    \STATE $\mathcal{L}_i \leftarrow \sum_{k\in K_1} x_i^k + c \cdot \sum_{k\in K_2} x_i^k$
    \IF{$(2 \mathcal{L}_i - \mu)(1 - 2p) > 0$}
    \STATE $x_i\leftarrow 1$
    \ELSIF{$(2 \mathcal{L}_i - \mu)(1 - 2p) < 0$}
    \STATE $x_i\leftarrow 0$
    \ELSE
    \STATE Sample $x_i$ uniformly on $\binarySet$
    \ENDIF
    \ENDFOR
    \STATE Evaluate $f(x)$
    \STATE $\widehat{p} \leftarrow \frac1{\mu} \cdot \sum_{k\in K_1} \frac{\norm{u^k}}{n} + \frac{c}{\mu} \cdot \sum_{k\in K_2} \frac{\norm{u^k}}{n}$
    \STATE $p \leftarrow p + \alpha \cdot (\widehat{p} - p)$
    \ENDLOOP
  \end{algorithmic}
\end{algorithm}

\section{Experiments}
\label{sec:experiments}

\subsection{Evolution of the mutation rate}

The first series of experiments\footnote{All experiments have been
  produced with the HNCO framework \cite{hnco}.} illustrates the
influence of the update rule and the hyperparameters on the evolution
of the mutation rate when it-EAs are applied to OneMax. It should be
noted that each plot in this subsection is the result of a single run
of a random search heuristics. In the following analysis, for each
plot, we concentrate on two indicators, the peak mutation rate and the
duration of the time spent above $p_{\min}$ which we will refer to as
the width of the plot. Then we study their relation with the changing
condition.

\begin{figure}
  \centering
  \includegraphics[width=\linewidth]{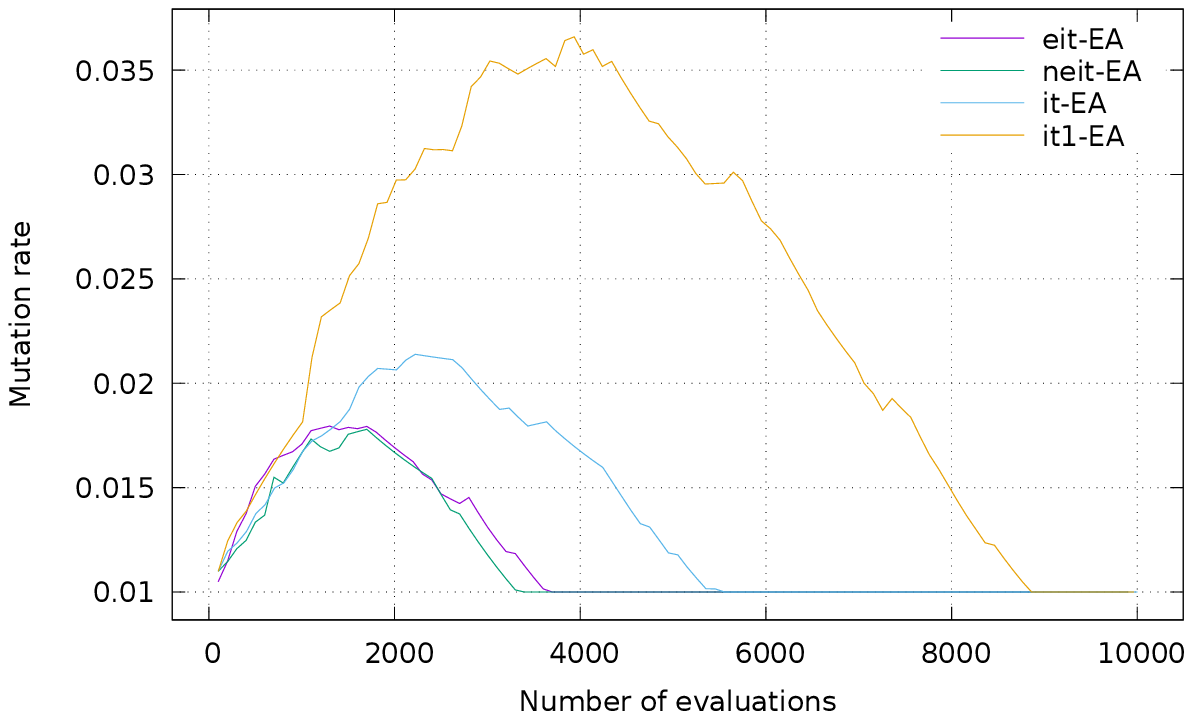}
  \caption{Influence of the update rule on the evolution of the
    mutation rate. The it-EAs are applied to OneMax with $n=100$,
    $\lambda=100$, $\mu = 1$, $p_{\min} = p_0 = 0.01$, and
    $\alpha = 0.05$.}
  \label{fig:influence_update_rule}
\end{figure}

Fig.~\ref{fig:influence_update_rule} shows that the update rule
influences the scale of the evolution of the mutation rate. The it1-EA
(local ML update) achieves the largest peak and width.

\begin{figure}
  \centering
  \includegraphics[width=\linewidth]{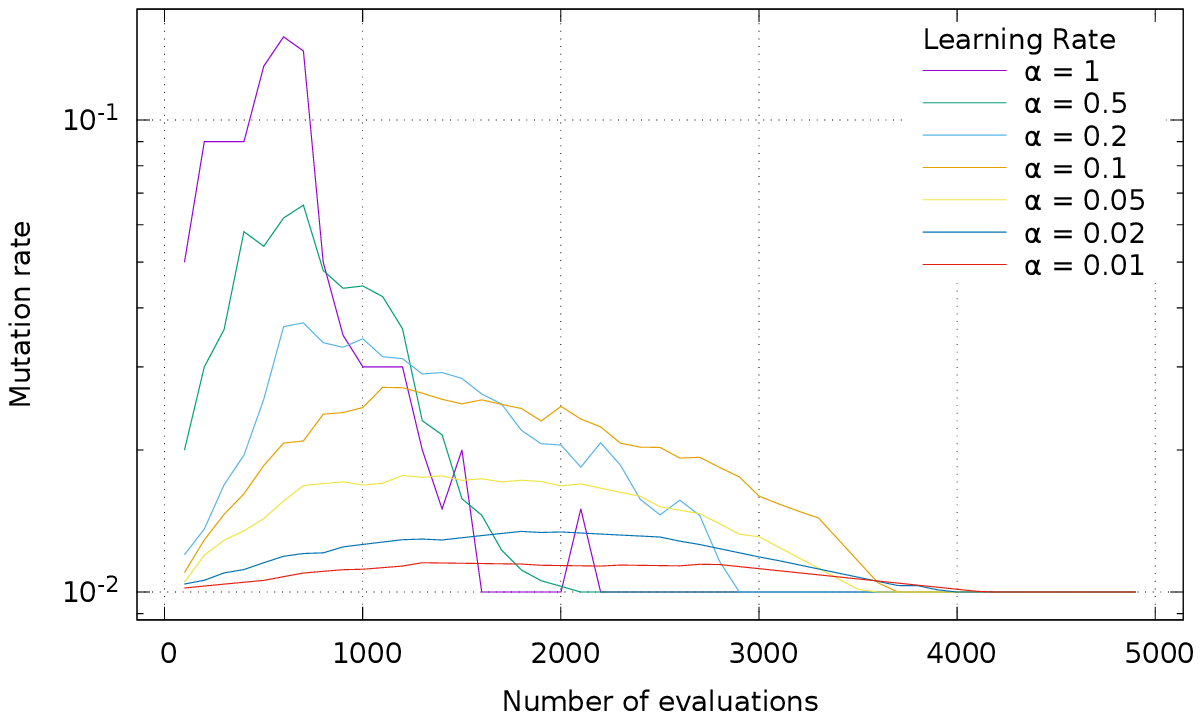}
  \caption{Influence of the learning rate on the evolution of the
    mutation rate. The eit-EA is applied to OneMax with $n=100$,
    $\lambda=100$, $\mu = 1$, and $p_{\min} = p_0 = 0.01$. The
    vertical axis uses a logarithmic scale.}
  \label{fig:influence_learning_rate}
\end{figure}

Fig.~\ref{fig:influence_learning_rate} shows the influence of the
learning rate. The peak increases with the learning rate whereas the
width decreases with it.

\begin{figure}
  \centering
  \includegraphics[width=\linewidth]{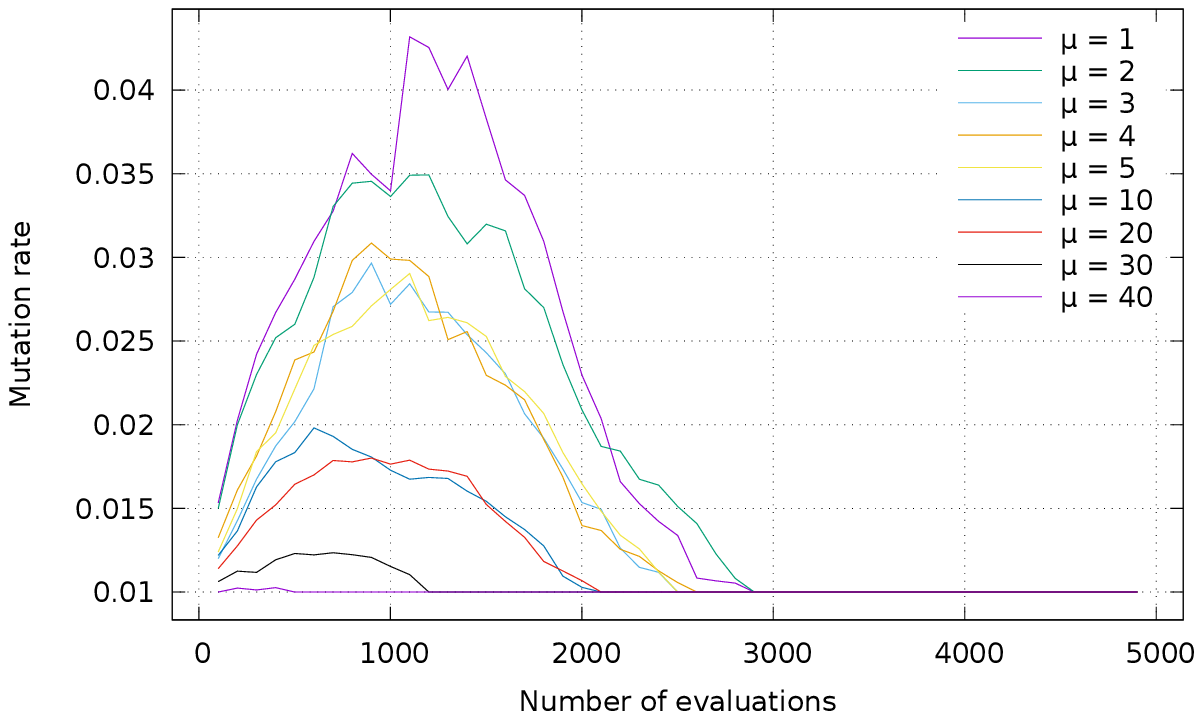}
  \caption{Influence of $\mu$ on the evolution of the mutation rate.
    The eit-EA is applied to OneMax with $n=100$, $\lambda=100$,
    $p_{\min} = p_0 = 0.01$, and $\alpha = 0.2$.}
  \label{fig:influence_mu}
\end{figure}

Fig.~\ref{fig:influence_mu} shows the influence of the number $\mu$ of
selected individuals. The value of $\mu$ mostly influences the peak
mutation rate. The smaller the value of $\mu$, the larger the peak
mutation rate.

\begin{figure}
  \centering
  \includegraphics[width=\linewidth]{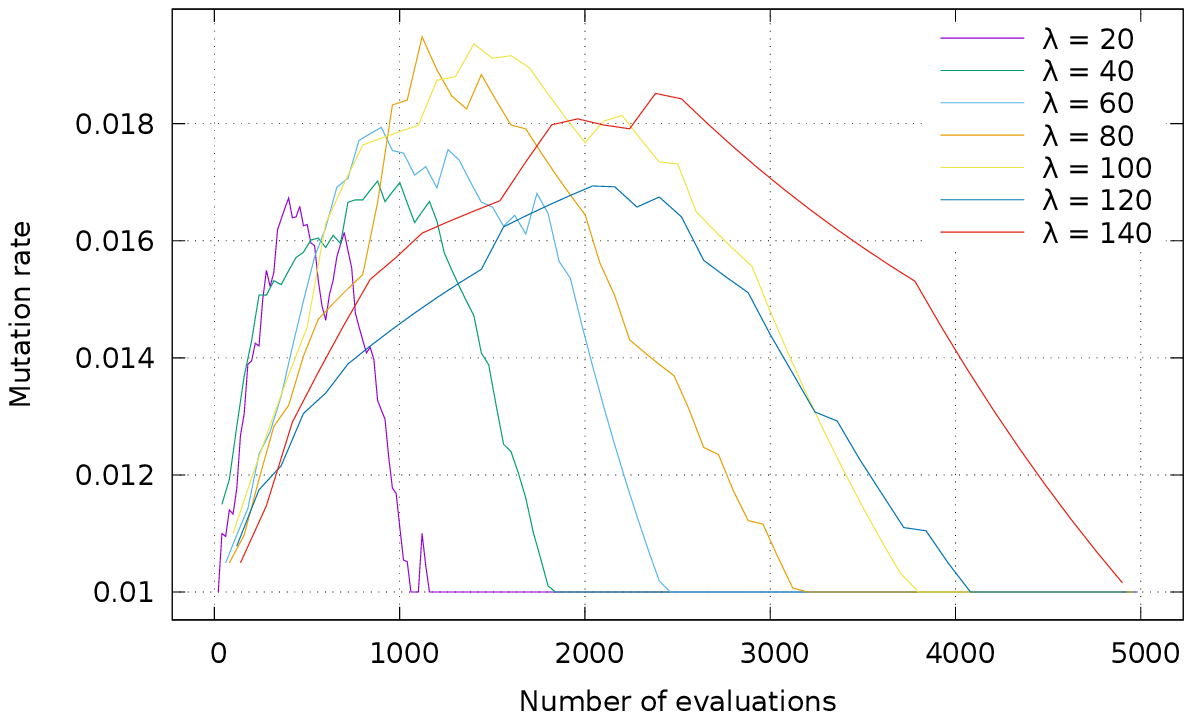}
  \caption{Influence of the population size $\lambda$ on the evolution
    of the mutation rate. The eit-EA is applied to OneMax with
    $n=100$, $\mu=1$, $p_{\min} = p_0 = 0.01$, and $\alpha = 0.05$.}
  \label{fig:influence_lambda}
\end{figure}

Fig.~\ref{fig:influence_lambda} shows the influence of the population
size $\lambda$. The value of $\lambda$ mostly influences the width of
the plot. The larger the value of $\lambda$, the larger the width.

\begin{figure}
  \centering
  \includegraphics[width=\linewidth]{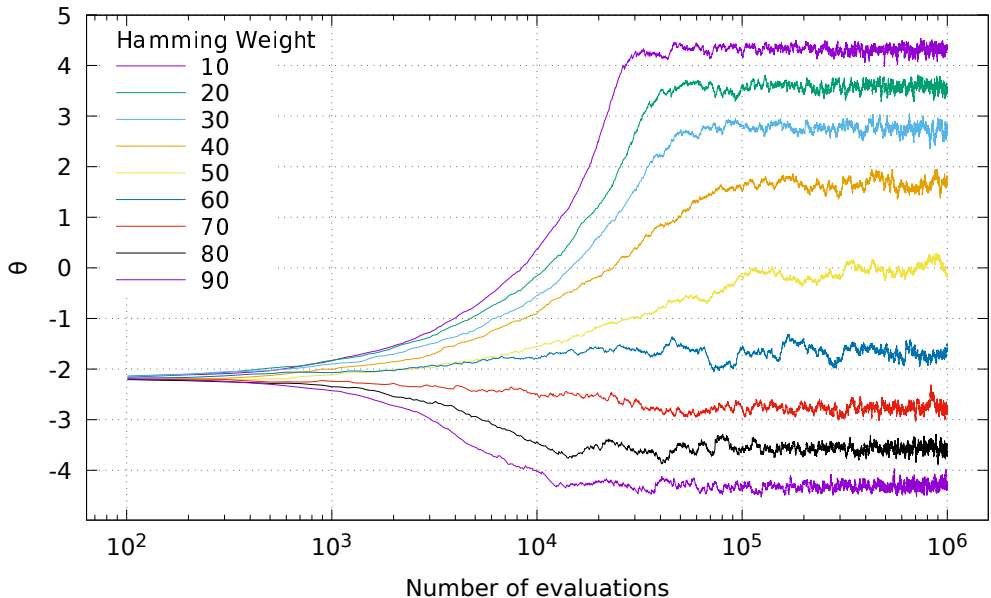}
  \caption{Evolution of $\theta = \logit(p)$ in static searches
    starting from random bit vectors of different Hamming weights. The
    it-EA is applied to OneMax with $n=100$, $\lambda=100$, $\mu=1$,
    $p_0 = 0.1$, $p_{\min} = 0$, $p_{\max} = 1$, and $\alpha = 0.05$.
    The horizontal axis uses a logarithmic scale.}
  \label{fig:static_search}
\end{figure}

In the next series of experiments, we study the convergence of the
mutation rate under the IGO update alone. To this end, we use
Algorithm~\ref{algo:it_ea} except that we do not update the center in
the loop. We apply this static search to OneMax starting from random
bit vectors with different Hamming weights. Instead of the mutation
rate $p$ itself, we represent $\theta = \logit(p)$ in order to better
visualize the range of asymptotic values. Fig.~\ref{fig:static_search}
shows that the mutation rate converges to asymptotic values closely
related to the Hamming weights of initial bit vectors. The larger the
Hamming weight, the smaller the asymptotic mutation rate. In
particular, for a 100-bit vector of Hamming weight 50, $\theta$
converges to 0 or, equivalently, $p$ converges to $1/2$. The symmetry
of Hamming weights about 50 is mapped onto the symmetry of $\theta$
values about 0.

\subsection{Fixed-target analysis}
\label{sec:fixed-targ-analys}

In the next series of experiments, we study the empirical runtime
(number of function evaluations until the maximum is found) of the
$(1+\lambda)$ EA, the 2-rate $(1+\lambda)$ EA \cite{doerr2017theone},
the eit-EA, and the neit-EA. The it-EA and it1-EA have proved slower
than the $(1+\lambda)$ EA and have not been included in the study. For
all algorithms, $\lambda = n/10$. For the eit-EA and the neit-EA,
$p_{\min} = p_0 = 1/n$ and $\alpha = 0.2$. For the 2-rate
$(1+\lambda)$ EA, $p_0 = 1/n$, which corresponds to
$r^\text{init} = 1$ in \cite{doerr2017theone}.

\begin{figure}
  \centering
  \includegraphics[width=\linewidth]{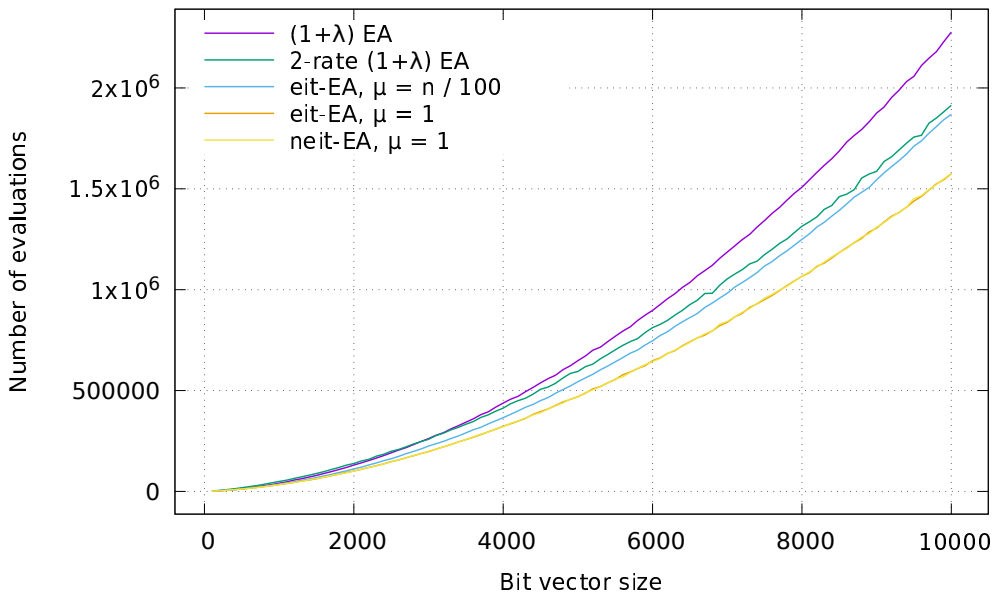}
  \caption{Mean runtime on OneMax as a function of $n$ (100 runs).}
  \label{fig:runtime_one_max_mean}
\end{figure}

Fig.~\ref{fig:runtime_one_max_mean} shows the mean runtime on OneMax
in dimensions up to $n = \np{10000}$ (100 runs). The eit-EA and
neit-EA with $\mu = 1$ have similar performance. Both are faster than
the eit-EA with $\mu = n / 100$ which, in turn, is faster than the
$(1+\lambda)$ EA. The 2-rate $(1+\lambda)$ EA is slower than the
eit-EA with $\mu = n / 100$ up to $n = \np{10000}$ but data suggests
that it is faster past a certain dimension. The ratio of the runtime
of the 2-rate $(1+\lambda)$ EA and that of the neit-EA with $\mu = 1$
is slowly decreasing with $n$ and is approximately equal to \np{1.22}
when $n = \np{10000}$. It is an open problem whether the eit-EA or the
neit-EA with $\mu = 1$ matches the optimal bound of the 2-rate
$(1+\lambda)$ EA.

\begin{figure}
  \centering
  \includegraphics[width=\linewidth]{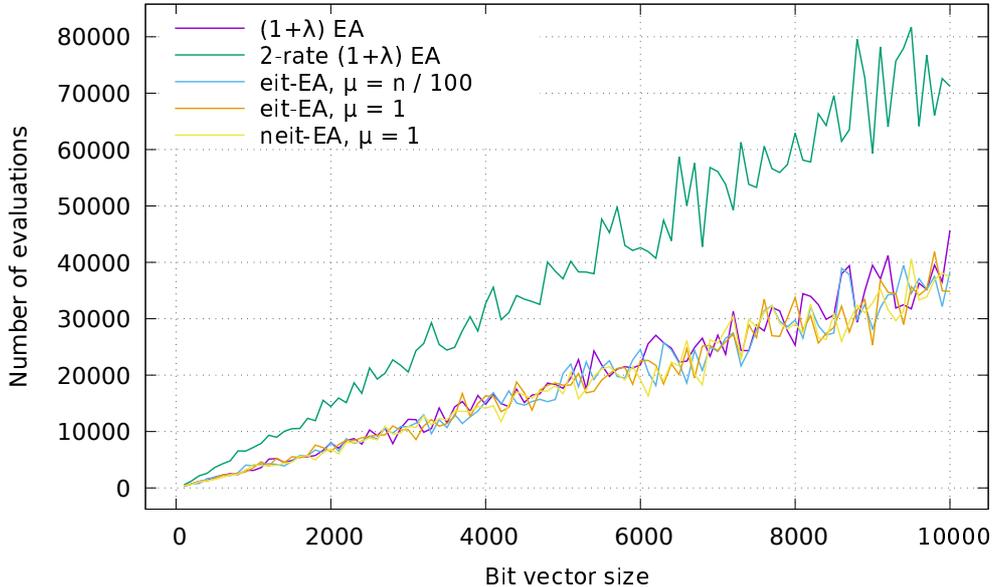}
  \caption{Standard deviation of the runtime on OneMax as a function
    of $n$ (100 runs).}
  \label{fig:runtime_one_max_stddev}
\end{figure}

Fig.~\ref{fig:runtime_one_max_stddev} shows the standard deviation of
the runtime on OneMax. It increases faster with the 2-rate
$(1+\lambda)$ EA than with the other algorithms in the study. One
explanation could be that, at each iteration, the 2-rate $(1+\lambda)$
EA multiplies or divides the mutation rate by two, each with
probability $1/4$. In contrast, in the it-EA, the change in the
mutation rate implied by Eq.~(\ref{eq:9}) is smooth.

\begin{figure}
  \centering
  \includegraphics[width=\linewidth]{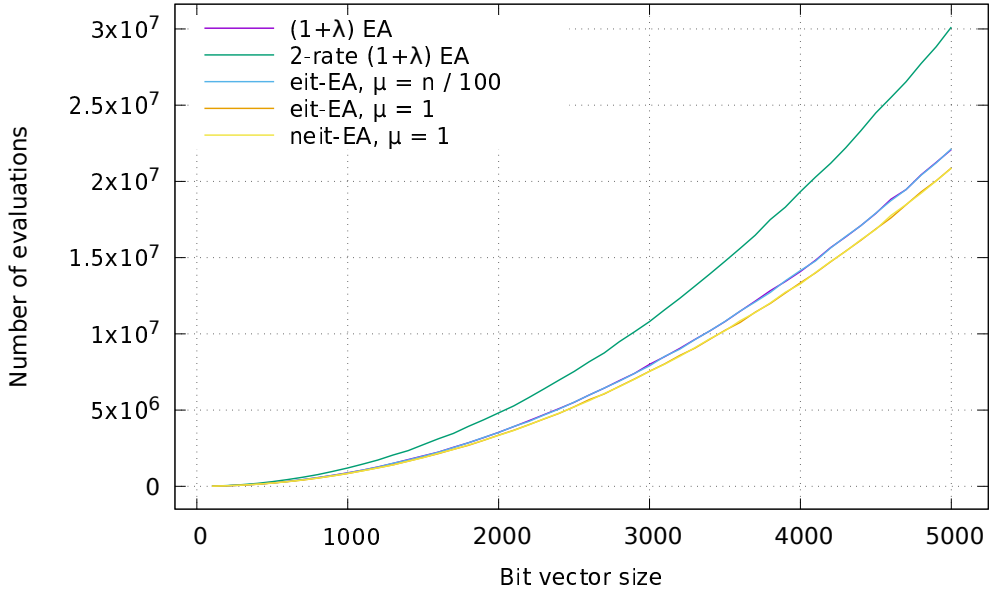}
  \caption{Mean runtime on LeadingOnes as a function of $n$ (100
    runs). The $(1+\lambda)$ EA and the eit-EA with $\mu = n / 100$
    have approximately the same runtime.}
  \label{fig:runtime_leading_ones_mean}
\end{figure}

Fig.~\ref{fig:runtime_leading_ones_mean} shows the mean runtime on
LeadingOnes in dimensions up to $n = \np{5000}$ (100 runs). The 2-rate
$(1+\lambda)$ EA is slower than the other algorithms in the study. The
eit-EA and neit-EA with $\mu = 1$ are marginally faster than the
$(1+\lambda)$ EA. The standard deviation of the runtime on LeadingOnes
increases faster with the 2-rate $(1+\lambda)$ EA than with the other
algorithms in the study (not shown).

\section{Conclusion}
\label{sec:conclusion}

We have presented a novel evolutionary algorithm on bit vectors, the
information theoretic evolutionary algorithm (it-EA). It relies on
standard bit mutation and rank-based selection and is completely
expressed in terms of information theory. The mutation rate is
controlled by means of information-geometric optimization. The IGO
update has a clear interpretation in terms of frequency of mutated
bits in selected individuals. The center of the search distribution is
updated by means of a maximum likelihood principle. The ML update can
be interpreted as a multi-parent cross-over operator. The it1-EA with
local ML update can be seen as the analogue of CMA-ES for bit vectors,
albeit with fewer parameters. We have also considered elitist (eit-EA)
and non elitist (neit-EA) selection for replacement.

A series of experiments involving the it-EAs has shown that the IGO
update for the mutation rate is effective. An empirical runtime
analysis has shown that, on OneMax in dimensions up to
$n = \np{10000}$, the eit-EA and the neit-EA with $\mu = 1$ are faster
than the $(1+\lambda)$ EA and even the 2-rate $(1+\lambda)$ EA. On
LeadingOnes in dimensions up to $n = \np{5000}$, they are marginally
faster than the $(1+\lambda)$ EA.

As future work, we plan to evaluate the performance of the it-EAs on
test functions other than OneMax and LeadingOnes. In particular, the
influence of hyperparameters on performance should be investigated.
The theoretical expected runtime of the eit-EA and the neit-EA on
OneMax is an interesting open question. Beyond the it-EA, the IGO
update (control of the mutation rate) and the ML update (multi-parent
cross-over) could find their way into other random search heuristics
on bit vectors. More generally, the relation between optimization and
learning could be deepened with the help of information theory.

\bibliographystyle{plain}

\bibliography{bibliography}

\end{document}